\documentclass[runningheads]{llncs}

\usepackage[T1]{fontenc} 
\usepackage[utf8]{inputenc}
\usepackage{amsmath}
\usepackage{newtxmath}
\usepackage{color}
\usepackage{verbatim}
\usepackage{bbm}
\usepackage[ruled,vlined]{algorithm2e}
\usepackage{graphicx}
\usepackage{graphics}
\usepackage{multirow}
\usepackage{tcolorbox}
\usepackage{enumitem} 
\usepackage[colorinlistoftodos,prependcaption]{todonotes}
\usepackage{xargs}
\usepackage{comment}
\usepackage{subfig}
\usepackage{pifont}
\usepackage{algorithmic}
\usepackage{colortbl}
\usepackage{xcolor}
\usepackage{makecell}

\usepackage{balance}

\usepackage{scalerel}
\usepackage{tikz}
\usetikzlibrary{svg.path}
\definecolor{orcidlogocol}{HTML}{A6CE39}
\tikzset{
  orcidlogo/.pic={
    \fill[orcidlogocol] svg{M256,128c0,70.7-57.3,128-128,128C57.3,256,0,198.7,0,128C0,57.3,57.3,0,128,0C198.7,0,256,57.3,256,128z};
    \fill[white] svg{M86.3,186.2H70.9V79.1h15.4v48.4V186.2z}
                 svg{M108.9,79.1h41.6c39.6,0,57,28.3,57,53.6c0,27.5-21.5,53.6-56.8,53.6h-41.8V79.1z M124.3,172.4h24.5c34.9,0,42.9-26.5,42.9-39.7c0-21.5-13.7-39.7-43.7-39.7h-23.7V172.4z}
                 svg{M88.7,56.8c0,5.5-4.5,10.1-10.1,10.1c-5.6,0-10.1-4.6-10.1-10.1c0-5.6,4.5-10.1,10.1-10.1C84.2,46.7,88.7,51.3,88.7,56.8z};
  }
}
\newcommand\orcidicon[1]{\href{https://orcid.org/#1}{\mbox{\scalerel*{
\begin{tikzpicture}[yscale=-1,transform shape]
\pic{orcidlogo};
\end{tikzpicture}
}{|}}}}

\usepackage{url}
\usepackage{hyperref}
\hypersetup{
    colorlinks=true,
    linkcolor=blue,
    citecolor=blue,
    filecolor=magenta,      
    urlcolor=cyan,
}

\usepackage{booktabs}
\usepackage{dsfont}
\usepackage{xspace}
\makeatletter
\DeclareRobustCommand\onedot{\futurelet\@let@token\@onedot}
\def\@onedot{\ifx\@let@token.\else.\null\fi\xspace}
\def\eg{\emph{e.g}\onedot} 
\def\ie{\emph{i.e}\onedot} 
 
\def\etc{\emph{etc}\onedot} 
 
\def\etal{\emph{et al}\onedot}
\makeatother

\def\argmin{\operatornamewithlimits{arg\,min}}

\newcommand{\qi}[1]{\textbf{\textcolor{violet}{qi: #1}}}

\definecolor{tabgray3}{rgb}{0.95,0.95,0.95}
\definecolor{tabgray2}{rgb}{0.90,0.90,0.90}
\definecolor{tabgray1}{rgb}{0.85,0.85,0.85}
\definecolor{top1}{rgb}{1.0, 0.6, 0.6} 
\definecolor{top2}{rgb}{0.94, 0.9, 0.55}
\definecolor{colorsteps}{rgb}{0.83, 0.83, 0.83}
\definecolor{top1-2}{rgb}{1.0, 0.78, 0.78} 
\definecolor{top1-3}{rgb}{1.0, 0.90, 0.90} 

\begin{document}
%

\title{\emph{DeepRepair}: Style-Guided Repairing for DNNs in the
Real-world Operational Environment}

%
%
%

\author{Bing Yu\inst{1}\and
Hua Qi\inst{1} \and
Qing Guo\inst{2}\and
Felix Juefei-Xu\inst{3}\and \\
Xiaofei Xie\inst{2} \and
Lei Ma\inst{1}\and
Jianjun Zhao\inst{1}
}

\authorrunning{B. Yu, H. Qi, Q. Guo, F. Juefei-Xu, X. Xie, L. Ma, and J. Zhao}

\titlerunning{\emph{DeepRepair}: Style-Guided Repairing for DNNs}

\institute{Kyushu University, Japan \and Nanyang Technological University, Singapore \and  Alibaba Group, USA
}

\markboth{}%
{Shell \MakeLowercase{\textit{et al.}}: Bare Demo of IEEEtran.cls for IEEE Communications Society Journals}
%



\maketitle

\begin{abstract}
Deep neural networks (DNNs) are being widely applied for various real-world applications across domains due to their high performance (\eg, high accuracy on image classification). Nevertheless, a well-trained DNN after deployment could oftentimes raise errors during practical use in the operational environment due to the mismatching between distributions of the training dataset and the potential unknown noise factors in the operational environment, \eg, weather, blur, noise \, etc. Hence, it poses a rather important problem for the DNNs' real-world applications: how to repair the deployed DNNs for correcting the failure samples (\ie, incorrect prediction) under the deployed operational environment while not harming their capability of handling normal or clean data. The number of failure samples we can collect in practice, caused by the noise factors in the operational environment, is often limited. Therefore, It is rather challenging how to repair more similar failures based on the limited failure samples we can collect.

In this paper, we propose a \textit{style-guided data augmentation for repairing DNN in the operational environment}. We propose a style transfer method to learn and introduce the unknown failure patterns within the failure data into the training data via data augmentation. Moreover, we further propose the \textit{clustering-based failure data generation} for much more effective style-guided data augmentation. We conduct a large-scale evaluation with fifteen degradation factors that may happen in the real world and compare with four state-of-the-art data augmentation methods and two DNN repairing methods, demonstrating that our method can significantly enhance the deployed DNNs on the corrupted data in the operational environment, and with even better accuracy on clean datasets. 
\end{abstract}



%

\section{Introduction}



Over the past few years, deep neural networks (DNNs) achieved competitive performance and have been widely deployed in the real world, with multifarious applications ranging from visual perception \cite{he2016deep,tan2019efficientnet,anonymous2021lambdanetworks} such as autonomous driving \cite{yu2020bdd100k,geiger2012we,chen2015deepdriving,caesar2020nuscenes}, face recognition \cite{schroff2015facenet,ijcai20_fakespotter,xu2015spartans,pr16_fkda} and generation \cite{karras2020analyzing,juefei2018rankgan}, object detection \cite{ren2015faster,bochkovskiy2020yolov4} and segmentation \cite{he2017mask,qiao2020detectors}, tracking \cite{xu2020train,neurips20_abba}, medical imaging analysis \cite{tian2020bias,cheng2020adversarial}, various low-level vision problems such as super-resolution \cite{maeda2020unpaired,dong2015image,abiantun2019ssr2}, denoising \cite{moran2020noisier2noise,mildenhall2018burst}, illumination correction \cite{guo2020zero,juefei2015encoding,juefei2015pokerface}, image inpainting \cite{zheng2019pluralistic,juefei2016fastfood}, deraining \cite{ren2019progressive,guo2020efficientderain}, dehazing \cite{mehta2020hidegan,ancuti2020ntire}, \etc., to automatic speech recognition and natural language processing \cite{devlin2018bert,brown2020language} applications such as machine translation \cite{fan2020beyond,schwenk2019ccmatrix,el2019massive}, hate speech detection \cite{mollas2020ethos,leite2020toxic}, sentiment analysis \cite{agarwal2011sentiment,bakshi2016opinion}, speech generation and voice assistant \cite{oord2016wavenet,shen2018natural}, \etc, to more recent intelligent agents for games \cite{vinyals2019grandmaster,berner2019dota,tian2017elf} and decision making for robotics applications \cite{andrychowicz2020learning}, \etc. 

A prominent advantage of DNN software, compared to traditional code-based software, is that DNN software can capitalize on the abundant real-world training data and high-capacity neural networks to learn the desired prediction or generative models, oftentimes surpassing human-level performance. One of the most milestone examples is what we have witnessed in the game of Go where AlphaGo \cite{silver2016mastering} and subsequent AlphaGo Zero \cite{silver2017mastering} as well as AlphaZero \cite{silver2018general} dominate the complicated game. Another example is in the realm of partial information game such as the Libratus \cite{brown2018superhuman,brown2017libratus}, which out-bluffed best human players in the poker game of no-limit Texas Hold'Em.


Despite the high performance and many successes among the aforementioned DNN-based applications, DNN software still suffers from reliability issues, \ie, oftentimes the well-trained DNNs that are deployed in the real world operational environment can behave erroneously and deviate from what they are designed for. This can be primarily caused by the gap between the real-world test data distribution $\mathcal{D_\mathrm{te}}$ in the operational environment and the distribution of the previously collected training data corpus $\mathcal{D_\mathrm{tr}}$. We say a DNN has a high level of generalizability when the DNN that is trained on data that follows $\mathcal{D_\mathrm{tr}}$ can perform well on test data that follows $\mathcal{D_\mathrm{te}}$. In other words, DNN software works the best when $\mathcal{D_\mathrm{tr}} \rightarrow\mathcal{D_\mathrm{te}}$. There are several aspects worth discussing as to why such a gap between $\mathcal{D_\mathrm{tr}}$ and $\mathcal{D_\mathrm{te}}$ exists. First of all, the amount of labeled training data for a supervised learning scenario is usually limited, either due to the cost of data collection and/or the cost of providing manual ground-truth labels. 
Second, even the DNNs are trained with very abundant training data, the training corpus can never cover all the real-world (potential noise) variations and perturbations. There will always be some corner cases and edge cases that are never seen by the DNNs during the training process, and sometimes, these corner cases can be the culprit for causing erroneous behavior of the deployed DNNs. Third, the trade-off between the DNN performance on the training set and the DNN generalizability on the unseen testing set is usually not well defined at a clear cut because, at training time, we usually do not observe the real-world testing data. Performing too well on the training data will even lead to overfitting and poor generalizability over testing data. It usually takes a separate validation set mimicking the real-world testing data as well as heuristics to determine the best trade-off strategy. However, issues remain because the validation set often does not represent the entirety of the real-world testing data of the upcoming deployed operational environment.

There have been some recent attempts to measure the quality and robustness of the deployed DNN software by discovering failure cases that cannot be predicted correctly, such as adversarial attack techniques~\cite{KurakinICLR2017,CW_2017_SSP,GoodfellowARXIV2014} and the deep learning testing techniques~\cite{pei2017deepxplore,tian2018deeptest,ma2018deepgauge,xie2019deephunter}, \etc. 
Under the real-world setting, the already-deployed DNNs could oftentimes raise errors during practical usage due to the mismatching between distributions of training dataset and real-world collected data that may cause failures by unknown real-world factors, \eg, weather, blur noise \etc. Hence, it poses a rather important problem for the DNNs' real-world applications: how to repair the deployed DNNs for the \emph{unknown} failure data while not harming their capability of handling normal or clean data. Recently, automatic repair\cite{gazzola2017automatic,monperrus2018automatic} has achieved promising progress in the traditional software, where faults in a program can be caused by specific lines of code. Thus, the repair can be performed to localize and repair the fault-triggering source code. Differently, DNN follows a data-driven programming paradigm and has no such explicit structure, which makes the repair of DNN still an open challenging research problem.

One of the most commonly used methods that can be used for DNN repair is to retrain the DNN with the failure samples. By discovering the failure samples (\eg, adversarial examples) and adding them to the training data, so that the failure samples could be repaired. 
Another technique is to directly modify the (hyper-)parameters of the model or the structure~\cite{maclaurin2015gradient,zhang2019apricot} based on the guidance of the failure methods. 
Some techniques~\cite{ma2018mode,eniser2019deepfault,borkar2019deepcorrect} are proposed to identify the buggy units of the DNN (\eg, the neurons) and then fix these errors by generating samples for retraining. The key challenge is that we usually can only collect a small number of failure cases, which represent limited failure patterns in the real-world environment. Thus, repair with direct retraining or evolving parameters may have low generalizability, \ie, they can only work well on the collected failure cases but may fail on other unseen failures. Although we can generate a large number of failure cases, it is still unknown whether these failure cases could represent diverse failure patterns. It is inefficient if many failures may be redundant, \ie, they share the same failure pattern.
Another challenge is that the repair must retain the performance and capability of the DNNs when handling data that can be predicted correctly before. How to achieve both goals simultaneously is a challenge itself.

Towards tackling the challenge, in this paper, we propose a novel method to
repair the DNN on image classification task. Our assumption is that we only collect a small number of failure data, based on which the DNN needs to be repaired. Specifically, we propose the \textit{style-guided data augmentation for DNN repairing} where a style transfer \cite{yoo2019ICCV,gatys2016image,luan2017deep,li2018closed} is proposed to introduce the unknown failure patterns (\eg, potential noise and combinations) within the failure data into the training data via data augmentation. Moreover, we propose the \textit{clustering-based failure data generation} for much more effective style-guided data augmentation. By generating diverse augmented data, we retrain the model such that the collected failure data and other similar failure data can be fixed.

We conduct a large-scale evaluation with fifteen possible degradation factors that may happen in the real world and compare with five state-of-the-art data augmentation methods and two DNN repairing methods,  demonstrating that our method is able to significantly enhance the deployed DNNs on the failure data with even better accuracy on clean datasets.

Overall, the contribution of this paper is summarized as follows:
\begin{itemize}
    \item We formulate the problem of DNN repairing based on a small limited number of failure cases and analyze the challenges.
    \item We originally propose a novel repair technique by style transfer based data augmentation for the DNNs in the real-world operational environment.
    \item We perform a large-scale evaluation of our proposed technique under fifteen potential degradation factors, comparing with six state-of-the-art methods as baselines.
\end{itemize}



The repair of DNNs, in general, can be rather challenging, this paper takes a special focus on repairing the potential issues that are caused by the noise patterns (that introduce data corruptions) from the operational environments, which we believe could be an important direction towards practical DNN application with high quality.

The rest of the paper is organized as the following: in Section~\ref{sec:background}, we go over the background with respect to DNN software repairing. In Section~\ref{sec:method}, we formally introduce our main technical contributions where we first formulate the problem of DNN repairing for some specified failure patterns. Then, we present the data augmentation-based solution for this problem and reveal the associated challenges, which are further addressed by the proposed style-guided data augmentation for DNN repairing. In Section~\ref{sec:settings}, various experiments are conducted to validate the proposed method as well as answer the posted four Research Questions. In Section~\ref{sec:results}, evaluation results are shown and analyzed in detail, and the Research Questions are fully addressed. Finally, we conclude our work in Section~\ref{sec:concl}.

\section{Related Work} \label{sec:background}


In this section, we first briefly discuss the recent progress on DNN software testing, where many research efforts have been spent on.
Then, we take more focus on the connections and differences between DNN software repair and traditional software repair, diving into several recent work on DNN software repair, with detailed discussion on their method formulation, advantages as well as disadvantages.



\subsection{DNN Software Testing}
To test the quality and detect defects in DNNs, various deep learning testing techniques have been recently proposed. Consider the fundamental difference between traditional software and DNN, some research focuses on the testing criteria \cite{pei2017deepxplore,ma2018deepgauge,ma2019deepct,kim2019guiding}. DeepXplore \cite{pei2017deepxplore} firstly proposes the neuron coverage towards measuring the adequacy of the test cases. A neuron is activated if its value is larger than a threshold. Neuron coverage measures the percentage of neurons that are activated. However, neuron coverage is coarse and several test inputs can achieve very high neuron coverage~\cite{sekhon2019towards,ncmislead}. DeepGauge \cite{ma2018deepgauge} proposes a set of multi-granularity testing criteria based on the neuron-based behaviors. For example, $k$-Multisection Neuron Coverage (KMNC) extends the neuron coverage that, it first profiles the training data and obtains the values of all training data. For each neuron, the range of all activation status is partitioned into $k$ intervals. Then KMNC measures the ratio of all covered sections of all neurons of a DNN by a set of test cases. Furthermore, the authors propose DeepCT~\cite{ma2019deepct} that considers the relationship of different neurons in a layer based on the combinatorial method of software testing. Kim \etal \cite{kim2019guiding} propose the coverage criteria that measure the surprise of the inputs, \ie, the distance between the inputs and the training data. The assumption is that surprising inputs introduce more diverse data such that more abnormal behaviors could be tested.

Based on the proposed testing criteria, a number of test generation techniques \cite{pei2017deepxplore,tian2018deeptest,xie2019deephunter,test:deepstellar,odena2018tensorfuzz} are proposed for detecting defects in DNNs. Specifically, DeepXplore \cite{pei2017deepxplore} and DeepTest \cite{tian2018deeptest} generate test cases based on the guidance of neuron coverage. In particular, DeepXplore adopts a differential testing method that determines whether an input is erroneous based on the cross-validation between multiple DNNs.   TensorFuzz \cite{odena2018tensorfuzz} and DeepHunter \cite{xie2019deephunter} propose the coverage-guided fuzzing techniques to test DNNs. DeepHunter integrates the coverage criteria from DeepGauge while TensorFuzz adopts the distance-based coverage criteria. While the aforementioned techniques mainly focus on feed-forward neurall network, DeepStellar \cite{test:deepstellar} proposes the coverage criteria and fuzzing technique for recurrent neuron network. The basic idea is to build an abstract model from the given RNN. A set of coverage criteria are then defined based on the abstract model.

While the existing fuzzing techniques mainly discover defects in the model, our method can be treated as a mitigation technique for repairing such potential defects. MODE~\cite{ma2018mode} proposes a DNN debugging technique on feedforward neural networks (FNN). Given an FNN, a feature map is constructed in each layer. By selecting one layer, MODE detects the buggy weights for the given failed inputs and fixes these bugs by generating new training samples. Different from our method, MODE mainly focuses on fixing the under-fitting or overfitting problems while our method focuses on the more general problem, \ie, the distribution shift between training samples and the real-world test samples. More comprehensive discussion on the deep learning testing can be referred to the recent survey~\cite{9000651}



\subsection{DNN Software Repairing}

In software engineering literature \cite{gazzola2017automatic}, there are two well-studied approaches for tackling program failures, \ie, software healing and software repairing. Software healing detects software failures in-the-field and makes amends by responding to the failures and restoring normal operations. The key is that the amendments are not deployed at the source code level, but rather deployed at runtime in order to mitigate runtime failures on the deployed applications. On the other hand, in software repairing, the amendments operations are mainly performed on the program source code level to remove fault that causes a failure. In this work, our fixing of the DNN software conceptually falls under the second category, software repair, where fixes are deployed on the source code level at testing and design time, as opposed to at runtime. 

One of the recent attempts for DNN software repair is Apricot \cite{zhang2019apricot} that aims at fixing DL model iteratively through a weight-adaptation method. The approach is based on two observations: (1) It will be increasingly more difficult for the DL model to retain a large proportion of its weights to capture all essential features as the number of inputs in the dataset $T_0$ grows. (2) Considering a pair of DNNs denoted as $D_0$ and $D_\mathrm{sub}$, where their training processes are identical except that model $D_0$ is trained on the entirety of the dataset $T_0$, while model $D_\mathrm{sub}$ is trained on a subset $S_0$ of $T_0$, and model $D_\mathrm{sub}$ is referred to as the reduced deep learning model (rDLM). Following this, the second observation is that each individual rDLM may not fully capture the essential features needed to classify one particular test case correctly, and if there is a set of rDLMs such that, on average, each one is more likely to classify the test case correctly, then the combined tendency of this set of rDLMs is more likely to classify this test case correctly. Based on such observation, the Apricot approach is carried out as the following. First, Apricot generates a set of rDLMs, given a pair of DNN model $D_0$ and a training set $T_0$, each training with a randomly selected subset of $T_0$. Second, for each failing test case $x$ of the input model $D_k$ at the $k$-th step, Apricot splits the set of rDLMs into 2 partitions: \texttt{IncorrectSubModels$(x)_k$} and \texttt{CorrectSubModels$(x)_k$}, where each rDLM in these two categories respectively classifies $x$ incorrectly and correctly. Third, Apricot takes the average of the corresponding weight assignments $w$ of the DLMs in both \texttt{IncorrectSubModels$(x)_k$} and \texttt{CorrectSubModels$(x)_k$} and further combine the two mean values into a single one. Last, Apricot can adjust the weight $w$ of $D_k$ accordingly and it continues to train the adjusted $D_k$ with $T_0$ to produce the next input model $D_{k+1}$ for the $(k+1)$-th iteration step. After all the iterations are carried out, the output of the procedure is a deep learning model with repaired weights. 

Another recent work SENSEI \cite{gao2020fuzz} has proposed a new algorithm to improve the robust generalization of DNNs using guided test generation techniques to address the data augmentation problem for the DNNs under natural environmental variations. The proposed data augmentation problem is cast as an optimization problem. In order to identify the worst natural environmental variant for the augmentation, each training input data goes through a space search based on the genetic algorithm (GA). The algorithm is carried out as the following: (1) at each iteration of the DNN training, for each training input data, the genetic algorithm explores a small set of variants of the input and selects the locally worst one for augmentation; (2) It then uses it as the GA seed for the search in the next epoch for gradually reaching the globally worst variant without needing to explicitly evaluate all the possible variants; (3) a further heuristic-based data selection technique named selective augmentation is used to substantially reduce the DNN training time under augmentation by allowing complete skipping of a training data at certain DNN training epochs based on the DNN's current robustness around that data point. The proposed method has shown effectiveness on various image classification datasets for improving DNN robustness through GA guided data augmentation. 

Sohn \etal have proposed a search-based automated program repair technique for DNNs called Arachne \cite{sohn2019search}, where it directly manipulates the neural network weights and searches the space of possible DNNs instead of retraining the DNNs. The search is guided by a specifically designed fitness function following Generate and Validate automatic program repair (APR) techniques. The Arachne follows traditional code-based automatic program repairs techniques with the following steps: (1) Arachne first adopts a fault localization technique by utilizing both positive and negative input data to retain correct behavior and to generate a patch and the representation of the patch is a set of real-numbered neural network weights; (2) Arachne then uses particle swarm optimization (PSO) as its search algorithm to update the selected neural network weights with values from the PSO candidate solution, and further calculates the fitness value based on the outcomes. The Arachne approach is evaluated on three image classification tasks using under-trained DNNs to induce unexpected behavior. The repairs produced by Arachne is focused more on targeted misbehavior with minimal perturbation on other behaviors. This is opposed to retraining-based DNN repairs that can alter the behavior significantly. 

On a separate thrust, Islam \etal \cite{islam2020repairing} have conducted a comprehensive study of bug repair patterns for five DNN libraries Caffe, Keras, Tensorflow, Theano, and Torch by using the DNN bugs dataset that consists of 415 bugs from Stack Overflow and 555 bugs from GitHub. The study has analyzed the following aspects of the fix patterns: (1) common bug fix patterns, (2) fix patterns across bug types, (3) fix patterns across libraries, (4) risk in fix, and (5) challenges in fixing DNN bugs. 

Different from these existing DNN repairing work, we take a special focus on DNN repairing for the incorrect behaviors introduced by the noise patterns during the real-world operational environment. We also take a novel style-transfer based approach. With the limited number of collected DNN failure examples from the operational environment, we perform style transfer to guide the data augmentation so that similar failure noise patterns in the operational environment would not cause the incorrect decision of the repaired DNN.



\section{Methodology}\label{sec:method}
In this section, we first formulate the problem of DNN repairing for some specified failure patterns (\ie, Sec.~\ref{subsec:dnnrepair}), which frequently happened and is of importance for the real-world applications in the operational environment. Then, we present the data augmentation-based method for this problem and reveal the challenges in Sec.~\ref{subsec:dataaugDNN}. To address the challenges, we propose a style-guided data augmentation method for DNN repairing in Sec.~\ref{subsec:styleaug} and introduce the detailed algorithm of our method in Sec.~\ref{subsec:algorithm}.

\subsection{DNN Repairing for Failure Patterns}\label{subsec:dnnrepair}
Following the general DNN training process, we can learn a DNN $\phi_\theta(\cdot)$ on a large-scale training dataset $\mathcal{D}^\text{t}$ where $\theta$ represents the network's parameters. We then deploy it for the real-world applications with the assumption that the training dataset has the same distribution as the data captured in the real world, which is denoted as $\mathcal{D}^\text{r}$.
Nevertheless, in practice, it is rather difficult to construct such a perfect training dataset and the pre-trained DNN might raise errors when the input data is corrupted by some unknown patterns caused by real-world noise factors, \eg, weather, blur, and other various kinds of noise, \etc, which are dependent on the task and operational environments where the DNN is deployed.

Currently, even with the state-of-the-art deep learning techniques for both data and network architectures, it is still difficult to train such a DNN that can address all real-world situations under various environments with high performance (\eg, high accuracy for image recognition task). Hence, it poses a pressing problem for the DNN's real-world applications: when a pre-trained DNN makes incorrect predictions on some data that may have specific failure patterns (\eg, noise), how could we enhance it without harming its performance on other normal data? 
For example, given a DNN $\phi_\theta(\cdot)$ offline trained on $\mathcal{D}^\text{t}$ and evaluated on a testing dataset $\mathcal{D}^\text{v}$ for the image classification task, we deploy it in the real world and can find that it usually misclassifies images corrupted by noises with some kind of pattern where we do not know its simulation way but we can collect some failure case examples and obtain $\mathcal{D}^\text{c}$. Then, the problem can be formulated as: with the  $\mathcal{D}^\text{c}$ and $\mathcal{D}^\text{t}$, how should we improve the accuracy of $\phi_\theta(\cdot)$ on the data having the similar failure patterns with $\mathcal{D}^\text{c}$ while not reducing the accuracy on $\mathcal{D}^\text{v}$? 
Specifically, we can represent it as follows:
%
\begin{align}\label{eq:basic}
\argmin_{\theta}\mathbb{E}_{(\mathbf{X},y)=\mathcal{T}(\{\mathcal{D}^\text{t},\mathcal{D}^\text{c}\})}~{J\Big(\phi_{\theta}(\mathbf{X}),y\Big)},
\end{align}
%
where $J(\cdot)$ denotes the task-related loss function and we use cross-entropy function for the image classification task. $(\mathbf{X},y)$ denotes an example and corresponding label from datasets.
$\mathcal{T}(\{\mathcal{D}^\text{t},\mathcal{D}^\text{c}\})$ defines the way of using the two datasets. 
For example, when we have $\{(\mathbf{X},y)=\mathcal{T}(\mathcal{D}^\text{c})|(\mathbf{X},y)\in\mathcal{D}^\text{c}\}$, it means that we only use the collected dataset $\mathcal{D}^\text{c}$ to fine-tune the DNN $\phi_\theta(\cdot)$. Obviously, since $\mathcal{D}^\text{c}$ is a small-scale dataset with limited failure patterns, above-mentioned way (\ie, only utilizing $\mathcal{D}^\text{c}$) would lead to an over-fit DNN that has poor performance on testing dataset $\mathcal{D}^\text{v}$.

\subsection{Data Augmentation-based DNN Repairing}\label{subsec:dataaugDNN}

To avoid the over-fitting issue, a simple solution is to employ data augmentation operations to extend the training dataset $\mathcal{D}^\text{t}$ and fine-tune the DNN.
The intuition behind this solution is that the diverse augmentation operations (or transformations that simulate the real-world noise patterns in the operational environment) could cover the unknown failure patterns. For example, we follow the state-of-the-art AugMix method \cite{hendrycks2020augmix} that can be represented as
\begin{align}\label{eq:augmix}
\argmin_{\theta} \mathbb{E}_{\substack{(\mathbf{X},\mathbf{X}_\text{aug1},\mathbf{X}_\text{aug2},y)\\=\mathcal{T}(\mathcal{D}^\text{t},\mathcal{O})}}
~J\Big(\phi_{\theta}(\mathbf{X}),y\Big) +  \lambda\text{JS}\Big((\phi_{\theta}(\mathbf{X}),\phi_{\theta}(\mathbf{X}_\text{aug1}),\phi_{\theta}(\mathbf{X}_\text{aug2})\Big),
\end{align}
where $\mathcal{T}(\mathcal{D}^\text{t},\mathcal{O})$ is to perform transformations on each sample $\mathbf{X}\in\mathcal{D}^\text{t}$ and return two augmented versions, \ie, $\mathbf{X}_\text{aug1}$ and $\mathbf{X}_\text{aug2}$,  with a series of operations sampled from the operation set $\mathcal{O}$. In the AugMix method, $\mathcal{O}$ contains rotation, equalization, translation, sharpness, \etc. Please find details in \cite{hendrycks2020augmix}. $\mathbf{X}$, $\mathbf{X}_\text{aug1}$, and $\mathbf{X}_\text{aug2}$ share the same label $y$.
$\text{JS}(\cdot)$ denotes the Jensen-Shannon diverse loss function that enforces the DNN predicting consistent  results for original and augmented examples.

However, since the failure patterns in the collected data $\mathcal{D}^\text{c}$ are diverse and produced by various real-world factors, the limited augmentation operations can hardly simulate those patterns, making the augmented data not diverse and many failure data is missed.
As a result, the fine-tuned DNNs are not repaired for handling the failures properly. 
To address this challenge, we originally propose the style-guided data augmentation where the failure patterns are employed to guide the data augmentation, for repairing DNNs more effectively.

\begin{figure*}[t] 
	\begin{center}
		\includegraphics[width=0.99\linewidth]{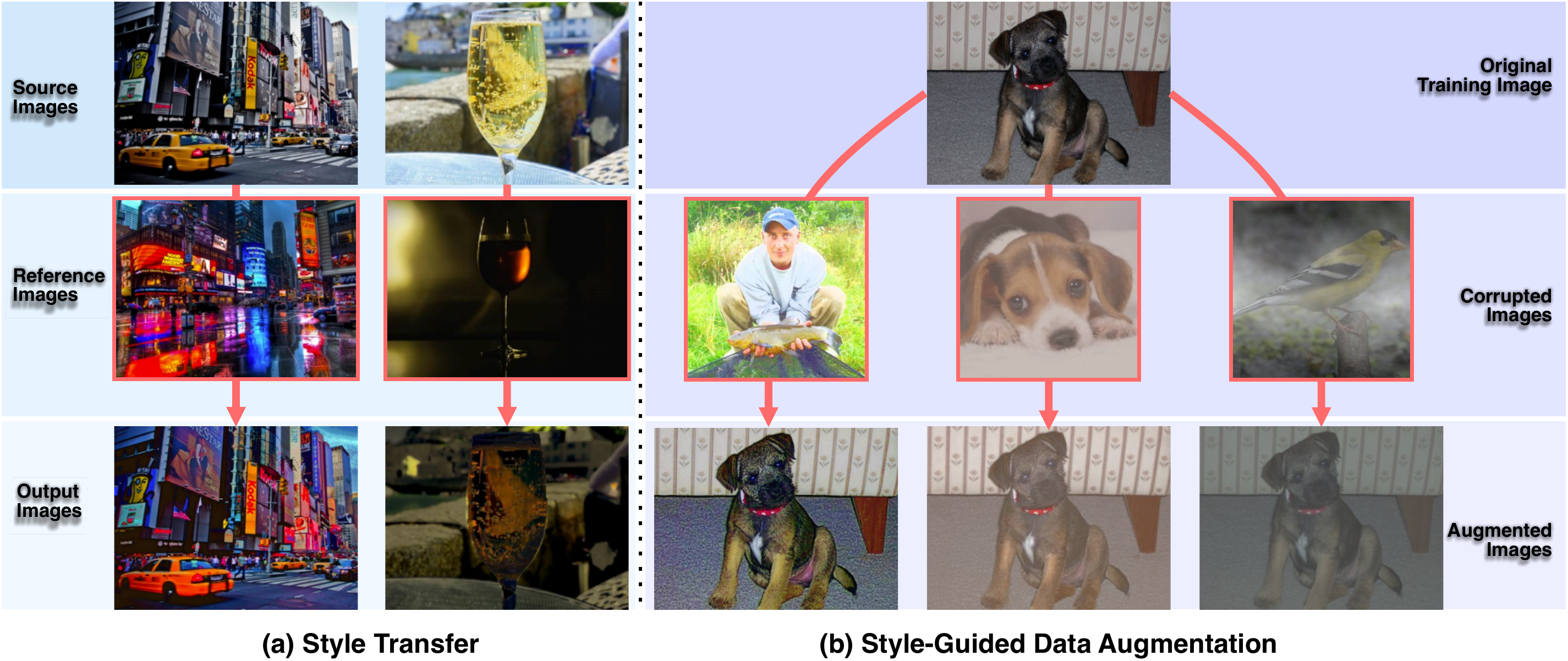}
	\end{center}
	\caption{\small{
		Two Examples (\ie, (a)) of Style transfer via \cite{yoo2019ICCV} and three examples (\ie, (b)) of our style-guided data augmentation for DNN Repairing.
	}}
	\label{fig:styletransfer}
\end{figure*}
%

\begin{algorithm}
\label{alg}
	\caption{\fontfamily{bch}\selectfont \footnotesize{Style-Guided DNN Repairing}}
	\label{algorithm}
	\LinesNumbered
	\KwIn{Training dataset $\mathcal{D}^\text{t}$, collected failure images $\mathcal{D}^\text{c}$, pre-trained DNN $\phi_\theta(\cdot)$, style transfer method,\ie, $\text{ST}(\cdot)$, augmentation operation set $\mathcal{O}$, and the pre-defined clustering number $N$.}
	\KwOut{Repaired DNN $\phi_{\bar{\theta}}(\cdot)$.}
	\colorbox{colorsteps}{\# \textit{Style-guided data augmentation}}\\
	\SetKwFunction{FMain}{\FuncSty{StyleAug}}
	\SetKwProg{Fn}{Function}{:}{}
	\Fn{\FMain{$\mathbf{X}$,$\mathcal{O}$}}{
	    Initialize $\mathbf{X}_\text{aug}$ with zeros and
	    sample mixing weights $(w_1,\ldots,w_M)~\sim~\text{Dirichlet}(\alpha,\ldots,\alpha)$\;
	    \For{$m=1,\ldots,M$}{
	    Sample the first operation by $\text{O}_1~\sim~\mathcal{O}$\;
	    Sample the $2$nd and $3$rd operations by $\{\text{O}_2,\text{O}_3\}~\sim~\mathcal{O}\diagdown\mathcal{O}^\text{c}$\;
	    Construct sequential operations: $\text{op}_{1}=\text{O}_{1}$,
	    $\text{op}_{12}=[\text{O}_1,\text{O}_2]$,
	    and $\text{op}_{123}=[\text{O}_1,\text{O}_2,\text{O}_{3}]$\;
	    Sample one operation, \ie, $\text{op}$, from $\{\text{op}_1,\text{op}_{12},\text{op}_{123}\}$\;
	    Conduct augmentation via $\text{op}$ and add it to $\mathbf{X}_\text{aug}$ with $\mathbf{X}_\text{aug}+=w_m\text{op}(\mathbf{X})$\;
	    }
	    Sample blending weight $w_0$ by $w_0~\sim~\text{Beta}(\alpha,\alpha)$\;
	    Blend with $\mathbf{X}$ by $\mathbf{X}_\text{mix}=w_0\cdot\mathbf{X}_\text{aug}+(1-w_0)\cdot\mathbf{X}$\;
        \textbf{return} $\mathbf{X}_\text{mix}$\; 
    }
	\colorbox{colorsteps}{\# \textit{Augmentation operation extension via the style transfer}}\\
	Perform K-means clustering on $\mathcal{D}^\text{c}$ with number $N$ \;
	Construct the sampling strategy $\mathcal{P}^\text{c}$ via Eq.~\ref{eq:sample_strategy} \;
	Identify $N$ operations $\mathcal{O}^\text{c}=\{\text{O}_i^\text{c}\}$ via Eq.~\eqref{eq:clsstypleop} \;
	Update the operation set $\mathcal{O}$ by adding  $\mathcal{O}^\text{c}$ to $\mathcal{O}$ \;
	\colorbox{colorsteps}{\# \textit{Data augmentation for DNN repairing}}\\
	\For{$j=1\ \mathrm{to}\ |\mathcal{D}^\text{t}|$}{
        Loading the $j$th image $\mathbf{X}$ from $\mathcal{D}^\text{t}$ \;
        Calculating two augmented images:
        $\mathbf{X}_{\text{aug1}}$ = StyleAug($\mathbf{X}$,$\mathcal{O}$)
        , and
        $\mathbf{X}_\text{aug2}$ = StyleAug($\mathbf{X}$,$\mathcal{O}$) \;
        Calculating the loss function:
        ${J\Big(\phi_{\theta}(\mathbf{X}),y\Big)+\lambda\text{JS}\Big((\phi_{\theta}(\mathbf{X}),\phi_{\theta}(\mathbf{X}_\text{aug1}),\phi_{\theta}(\mathbf{X}_\text{aug2})\Big)}$;
        Updating the parameters $\theta$ of $\phi_\theta(\cdot)$ \;
}
\end{algorithm}

\subsection{Style-Guided Data Augmentation for DNN Repairing}\label{subsec:styleaug}
Following the objective function in Sec.~\ref{subsec:dataaugDNN}, we focus on adding novel data augmentation operations to $\mathcal{O}$ for DNN repairing with the guidance of collected failure examples, \ie, corrupted data $\mathcal{D}^\text{c}$.
To this end, we propose the very first style transfer-based data augmentation operations.
Style transfer is to map an image to a new one having similar style with a given reference image. As shown in Fig.~\ref{fig:styletransfer}~(a), source images can be transferred to very similar styles with the given style images while the original details are all preserved. 
Intuitively, we could employ style transfer as novel data augmentation operations by regarding the corrupted data as reference images.
Specifically, we represent the new operations as 
%
\begin{align}\label{eq:stypleop}
\text{O}_i^\text{c}(\mathbf{X})=\text{ST}(\mathbf{X},\mathbf{X}^i),~ \text{with}~\mathbf{X}^i=\text{Sample}(\mathcal{D}^\text{c},\mathcal{P}),
\end{align}
%
%
where $\mathbf{X}$ is the example that needs to be augmented, the image $\mathbf{X}_i$ is one of the collected failure images, and the $\text{ST}(\cdot)$ denotes the style transfer method.
Here, we adopt the style transfer method proposed by \cite{yoo2019ICCV}.
The $\text{Sample}(\mathcal{D}^\text{c},\mathcal{P})$ denotes the sampling strategy based on $\mathcal{P}$. The notation $\mathcal{P}=\{P^i\}$ defines the sampling probability of all samples $\{\mathbf{X}^i\}$ and we preliminarily use the uniform sampling strategy with $\{P^i=\frac{1}{|\mathcal{D}^\text{c}|}\}$. 
As shown in Fig~\ref{fig:styletransfer} (b), the example with a dog is augmented according to three reference images that are predicted erroneously by over brightness, low contrast, and fog, and these failure patterns are successfully included into the augmented images.
We then add all style transfer-based operations, \ie, $\mathcal{O}^\text{c}=\{\text{O}_i^\text{c}|i=[1,\ldots,|\mathcal{D}^\text{c}|]\}$, to the operation set $\mathcal{O}$ and conduct the DNN repairing by fine-tuning DNNs via Eq.~\eqref{eq:augmix}.
However, there is still another challenge for DNN repairing. In particular, the failure images in $\mathcal{D}^\text{c}$ are diverse. Thus, for an image  $\mathbf{X}$, it is time-consuming to select all failure images as the reference. Thus, it is difficult to select which images in $\mathcal{D}^\text{c}$ should be the references. 

To alleviate this issue, we further implement the clustering-based reference image generation where the sampling probability of each sample is determined by their distance to the clustering center.
In particular, we first perform k-means clustering on the $\mathcal{D}^\text{c}$ with the number of clusters $N$ and get $N$ subsets denoted as $\{\mathcal{D}^{\text{c}_i}|i=[1,N]\}$ with their clustering centers being $\mathcal{C}^\text{c}=\{\mathbf{X}_\text{cls}^i|i=[1,N]\}$. Then, for the $i$th clustering set (\ie, $\mathcal{D}^{\text{c}_i}$), we calculate $L_2$ distance between samples in $\mathcal{D}^{\text{c}_i}$ and the clustering center $\mathbf{X}_\text{cls}^i$, which is represented as $\{d^i_j=\|\mathbf{X}_j^i-\mathbf{X}_\text{cls}^i\|_2|\mathbf{X}_j^i\in\mathcal{D}^{\text{c}_i}\}$ and define the sampling probability of $\mathbf{X}_j^i$ as 
%
\begin{align}\label{eq:sample_strategy}
P_j^i = \frac{1}{N}(1-\frac{d_j^i}{\sum_{k=1}^{|\mathcal{D}^{\text{c}_i}|}d_k^i}), 
\end{align}
%
where $|\mathcal{D}^{\text{c}_i}|$ is the size of $\mathcal{D}^{\text{c}_i}$ and $P_j^i$ is the probability of  $\mathbf{X}_j^i$ to be sampled for guiding data augmentation. Intuitively, the sample near to clustering center has higher probability to be selected. 
We define $\mathcal{P}^{c}=\{P_j^i \}$ as the clustering-guided sampling strategy and reformulate  Eq.~\eqref{eq:stypleop} as
%
\begin{align}\label{eq:clsstypleop}
\text{O}_i^\text{c}(\mathbf{X})=\text{ST}(\mathbf{X},\mathbf{X}^i),~\text{with}~\mathbf{X}^i=\text{Sample}(\mathcal{D}^\text{c},\mathcal{P}^\text{c}).
\end{align}
%
%

%
\subsection{The DNN Repairing Algorithm}\label{subsec:algorithm}
Algorithm~\ref{algorithm} presents the whole process of our proposed method, which can be roughly decomposed into two stages: \ding{182} extending the augmentation operation set $\mathcal{O}$ via the style transfer based on collected failure cases (Line 2-12); \ding{183} 
conducting the data-augmentation for DNN repairing through the style-guided data augmentation (Line 14-22). We first introduce the style-guided data augmentation, \ie, $\FuncSty{StyleAug}(\cdot)$ in Algorithm~\ref{algorithm}. Intuitively, given an input image $\mathbf{X}$, we obtain $M$ augmentations via the sequential operations sampled from the set $\mathcal{O}$ and then mix them up together with weights from Dirichlet and Beta distributions. We use these two distributions since their capability for data augmentation has been validated in \cite{baseline:augmix} and \cite{baseline:mixup}.

The style-guided augmentation is based on the framework AugMix~\cite{baseline:augmix}. 
More specifically, we obtain $M$ weights via Dirichlet distribution with $\alpha$ as the parameter (\ie, the $4$th line in Algorithm~\ref{algorithm}), and then perform $M$ augmentations (\ie, line $5$ to line $10$ in Algorithm~\ref{algorithm}). For each augmentation, we first sample the first operation (\ie, $\text{O}_1$) from $\mathcal{O}$, which might be the style transfer-based operations in $\mathcal{O}^\text{c}$ or other general operations in $\mathcal{O}\diagdown\mathcal{O}^\text{c}$ (\eg, rotation, translation). Then, we sample the second and third operations from $\mathcal{O}\diagdown\mathcal{O}^\text{c}$ and get $\{\text{O}_2,\text{O}_3\}$. All sampled operations are sequentially composed, resulting in three sequential operations, one of which is selected for the final augmentation.
Above process is shown from line $6$ to $10$ in the Algorithm~\ref{algorithm} with the following principles : \ding{182} the first operation could be based on style transfer, using to embed the failure pattern in collected failure examples (\ie, $\mathcal{D}^\text{c}$) into the example $\mathbf{X}$. Note that, we also allow the first operation to be other general operations to avoid the risk of overfitting on the specific pattern in $\mathcal{D}^\text{c}$. \ding{183} the second and third operations are limited to be general operations (\ie, $\mathcal{O}\diagdown\mathcal{O}^\text{c}$), simulating the transformations on the style transferred example, \ie, by translation, rotation, equalization, sharpness, \etc.  Besides, since style transfer \cite{yoo2019ICCV} can be much slower than the general operations, thus $\text{O}_2$ and $\text{O}_3$ also avoid great time cost  with only style-based augmentation.

During training (Line 19-22), we firstly collect all transformation set including the style-based transformation and the general transformation (Line 17). For each training data $\mathbf{X}$, we calculate two augmented images $\mathbf{X}_{aug1}$ and $\mathbf{X}_{aug2}$, the model should have the similar predictions on $\mathbf{X}$, $\mathbf{X}_{aug1}$ and $\mathbf{X}_{aug2}$ (Line 22).

\section{EXPERIMENTAL DESIGN AND SETTINGS}\label{sec:settings}


%
We conduct a large-scale experiments to validate the proposed methods and try to investigate the following research questions:
\begin{itemize}[leftmargin=*]
    \item \textbf{RQ1.} What are the effects of different failure patterns (\ie, failure types) on DNN operational inference?
    \item \textbf{RQ2.} Does our proposed method outperform state-of-the-art (SOTA) data-driven repairing methods on the examples with specific failure patterns?
    \item \textbf{RQ3.} Does our proposed method harm the robustness to other failure patterns and clean data?
    \item \textbf{RQ4.} Do our proposed components all contribute the final performance (\ie, accuracy on different failure patterns)?
\end{itemize}

\subsection{Experimental Setups}\label{sec:setup}
To answer the above four questions, we consider the following setups on dataset, DNN architectures, related hyper-parameters, \etc.
%
%

{\bf Datasets.}
%
Following the formulations in Sec.~\ref{subsec:dnnrepair}, 
we employ the training dataset CIFAR-10 as the $\mathcal{D}^\text{t}$ in Sec.~\ref{subsec:dnnrepair} and extend its testing dataset, \ie, $\mathcal{D}^\text{v}$, via various failure types to validate our method.
Specifically, we first train a DNN (\ie, $\phi_\theta$) on the CIFAR-10's training dataset and select $15$ failure types (\ie, 15 different failure patterns) \cite{hendrycks2019robustness} \footnote{The $15$ failure patterns are: Gaussian noise, shot noise, impulse noise, defocus blur, glass blur, motion blur, zoom, snow, frost, fog, brightness, contrast, elastic transform, pixelate, and JPEG compression.} as the specific patterns that commonly exist in the real world and could make the accuracy of the DNN reduce significantly. 
Note that each of the 15 failure datasets contains five different severity levels of corrupted images. 
Then, following the setting in \cite{hendrycks2019robustness}, we apply the $15$ the potential failure patterns to $\mathcal{D}^\text{v}$, respectively, and generate $15$ new testing datasets that are denoted as $\{\mathcal{D}^{\text{v}_k}|k\in[1,\ldots,15]\}$, each of which is five time larger than $\mathcal{D}^{\text{v}_k}$ and has $50,000$ images since we consider five different severity levels for each pattern. 
Note that, some of the $50,000$ images may not be failures on the DNN. Thus, for the $k$th failure pattern, we evaluate the pre-trained DNN on all generated images $\mathcal{D}^{\text{v}_k}$ and identify the failure cases.
From the failure cases, we randomly select 1,000 failure cases as the dataset $\mathcal{D}^{\text{c}_k}$ while the residual failure cases form the dataset $\mathcal{D}^{\text{e}_k}$ (\ie, unknown failure cases).
Table~\ref{tab:datasets} shows the detailed number of each failure type for different models, where Column  $\mathcal{D}^{\text{c}_k}+\mathcal{D}^{\text{e}_k}$ represents the number of all failure cases.
Note that, the number of $\mathcal{D}^{\text{e}_k}$ is greater than that of $\mathcal{D}^{\text{c}_k}$. 
%
Intuitively, in terms of the $k$th failure type, our method is to repair the DNN $\phi_\theta(\cdot)$ to make it achieve high accuracy on the corresponding failure dataset $\mathcal{D}^{\text{e}_k}$ with the guidance of $\mathcal{D}^{\text{c}_k}$ while not harming the accuracy on other failure and the original testing datasets. Hence, we use the accuracy of repaired DNN on $\{\mathcal{D}^{\text{e}_k}|k\in[1,\ldots,k]\}$ to evaluate the performance of DNN repairing methods.


{\bf DNN architectures.}
%
We select three different state-of-the-art architectures (\ie, all convolution network (AllConvNet) \cite{allconv}, DenseNet \cite{densenet}, and Wide Residual Net (WideResNet)) \cite{wideresnet} as the DNNs to be repaired. For each architecture, we first pre-train them with original CIFAR-10's training set (\ie, $\mathcal{D}^\text{t}$), and the model with the highest accuracy in testing set (\ie, $\mathcal{D}^\text{v}$) will be saved. 

{\bf Hyper-parameters}
In terms of the training setup, we employ stochastic gradient descent (SGD) optimizer with batch-size of 128, learning rate of 0.1 and decay of 0.0005. Jensen-Shannon divergence will be used as loss function. The max epoch number is 500, and the training will stop if validation loss does not decrease in 10 epochs. As for AugMix, mixture width is set as 3, and mixture depth is randomly changed between 1 to 3. We set 9 base operations (\ie,autocontrast, equalize, posterize, rotate, solarize, shear-x, shear-y, translate-x and translate-y) to the operation set $\mathcal{O}$ in the algorithm~\ref{algorithm} with additional operations proposed in Sec.~\ref{subsec:algorithm}. We set the number of clusters, \ie, $N$, in Sec.~\ref{subsec:styleaug} as $5$.



{\bf Other configurations.}
We use PyTorch as the platform and all the experiments were preformed on a server with the Ubuntu 16.04 system with 12-core 3.6GHz Xeon CPU, 126GB RAM and 2 NVIDIA GeForce RTX 2080 Ti 12G GPUs.

{\bf Baselines.} 
We consider two kinds of baselines. The first set of baselines is the general data augmentation methods, \ie, AugMix \cite{baseline:augmix}, CutMix \cite{baseline:cutmix}, CutOut \cite{baseline:cutout}, and MixUp \cite{baseline:mixup}. The second set of baselines is recently proposed DNN repairing methods, \ie, SENSEI \cite{baseline:sensei} and Few-Shot \cite{Ren2020}. Specifically, for the first four data augmentation methods, we perform DNN repairing by using them to fine-tune the DNNs with the original training dataset (\ie, $\mathcal{D}^\text{t}$) and 1000 collected failure cases (\ie, $\mathcal{D}^{\text{c}_k}$). For the SENSEI and Few-Shot methods, we also employ the failure cases during repairing for a fair comparison.

\begin{table*}
    \centering
    \tiny
    \resizebox{1\linewidth}{!}{
    \begin{tabular}{c|l|ccc|ccc|ccc}
    \toprule
        \multicolumn{2}{c|}{\multirow{2}{*}{Dataset}} & \multicolumn{3}{c|}{WideResNet} & \multicolumn{3}{c|}{DenseNet} & \multicolumn{3}{c}{AllConvNet} \\
        \multicolumn{2}{c|}{} & $\mathcal{D}^{\text{c}_k}+\mathcal{D}^{\text{e}_k}$ & $\mathcal{D}^{\text{c}_k}$ & $\mathcal{D}^{\text{e}_k}$ & $\mathcal{D}^{\text{c}_k}+\mathcal{D}^{\text{e}_k}$ & $\mathcal{D}^{\text{c}_k}$ & $\mathcal{D}^{\text{e}_k}$ & $\mathcal{D}^{\text{c}_k}+\mathcal{D}^{\text{e}_k}$ & $\mathcal{D}^{\text{c}_k}$ & $\mathcal{D}^{\text{e}_k}$ \\
    \midrule
        \multirow{15}{*}{\rotatebox{90}{CIFAR-10-C}} 
        & Gaussian noise (GN)   & 24026 & 1000 & 23026 & 24458 & 1000 & 23458 & 28218 & 1000 & 27218 \\
        & Shot noise (SN)        & 19701 & 1000 & 18701 & 19727 & 1000 & 18727 & 22561 & 1000 & 21561 \\
        & Impulse noise (IN)     & 19591 & 1000 & 18591 & 20189 & 1000 & 19189 & 22074 & 1000 & 21074 \\
        & Defocus blur (DB)     & 10398 & 1000 &  9398 & 11465 & 1000 & 10465 & 11497 & 1000 & 10497 \\
        & Glass blur (GB) & 28162 & 1000 & 27162 & 27844 & 1000 & 26844 & 25588 & 1000 & 24588 \\
        & Motion blur (MB) & 14055 & 1000 & 13055 & 13937 & 1000 & 12937 & 14058 & 1000 & 13058 \\
        & Zoom (ZM)       & 14625 & 1000 & 13625 & 14895 & 1000 & 13895 & 14247 & 1000 & 13247 \\
        & Snow (SW)            & 12907 & 1000 & 11907 & 13613 & 1000 & 12613 & 12359 & 1000 & 11359 \\
        & Frost (FT)          & 14840 & 1000 & 13840 & 15082 & 1000 & 14082 & 14856 & 1000 & 13856 \\
        & Fog (FG)            &  8222 & 1000 &  7222 &  8812 & 1000 &  7812 &  9207 & 1000 &  8207 \\
        & Brightness (BS)       &  5515 & 1000 &  4515 &  5735 & 1000 &  4735 &  4614 & 1000 &  3614 \\
        & Contrast (CT)        & 13466 & 1000 & 12466 & 13067 & 1000 & 12067 & 17320 & 1000 & 16320 \\
        & Elastic Transform (ET) & 11490 & 1000 & 10490 & 11569 & 1000 & 10569 &  9359 & 1000 &  8359 \\
        & Pixelate  (PIX)     & 15241 & 1000 & 14241 & 17865 & 1000 & 16865 & 13901 & 1000 & 12901 \\
        & JPEG Compression (JPEG) & 12864 & 1000 & 11864 & 12451 & 1000 & 11451 & 10377 & 1000 &  9377 \\
    \bottomrule
    \end{tabular}
    }
    \caption{The number of all failure cases (\ie, $\mathcal{D}^{\text{c}_k}+\mathcal{D}^{\text{e}_k}$), collected failure cases for repairing guidance (\ie, $\mathcal{D}^{\text{c}_k}$), and residual cases for repairing evaluation (\ie, $\mathcal{D}^{\text{e}_k}$) of three pre-trained DNNs, \ie, AllConvNet, DenseNet, and WideResNet.}
    \label{tab:datasets}
\end{table*}
    
\section{EXPERIMENTAL RESULTS}\label{sec:results}

\subsection{RQ1. What are the effects of different failure patterns (\ie, failure types) on DNN operational inference?}
\label{subsec:rq1}

\begin{figure}[t]
    \centering
    \includegraphics[width=\linewidth]{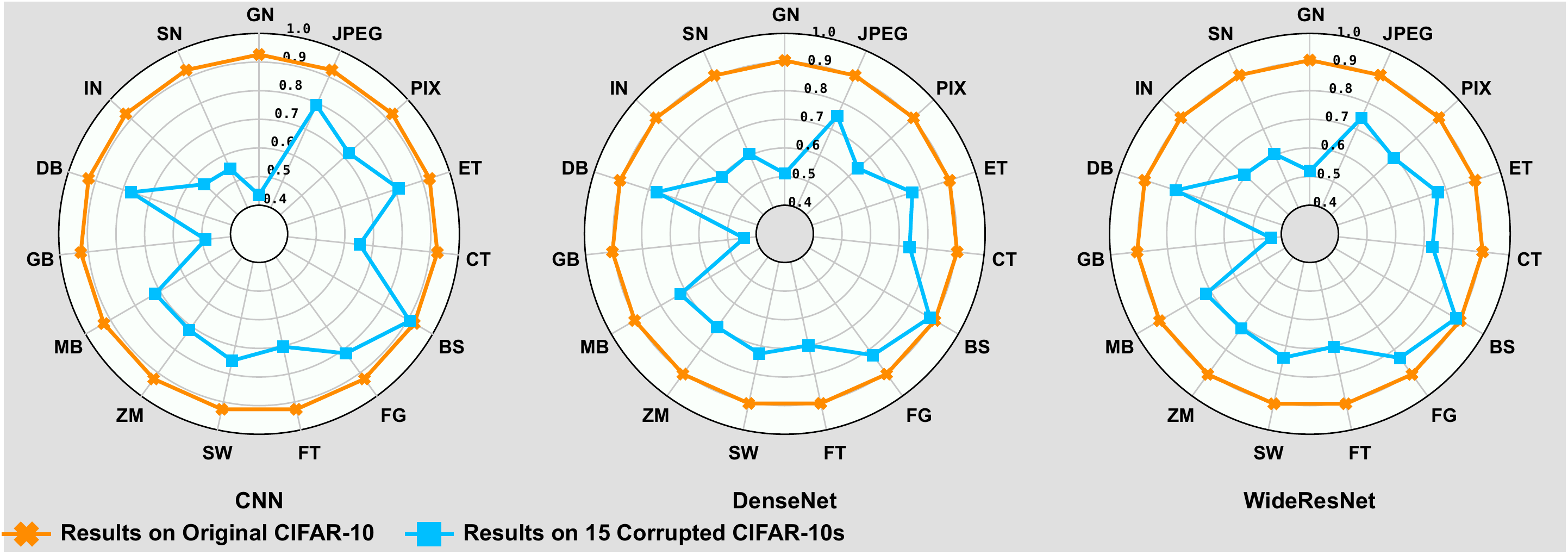}
    \caption{Accuracy of the original DNNs on CIFAR-10's testing dataset (\ie, $\mathcal{D}^\text{v}$)  and on 15 corrupted testing datasets (\ie, $\{\mathcal{D}^{\text{v}_k}|k=1,\ldots,15\}$).}
    \label{fig:rq1_polar}
\end{figure}

In this part, we train three DNNs, \ie, AllConvNet, DenseNet, and WideResNet, on the CIFAR-10’s training dataset (\ie, $\mathcal{D}^\text{t}$) and evaluate them on the original testing dataset (\ie, $\mathcal{D}^\text{v}$) and $15$ extended testing datasets (\ie, $\{\mathcal{D}^{\text{v}_k}|k=1,\ldots,15\}$). The evaluation results are summarized and presented in Fig.~\ref{fig:rq1_polar} and we have the following observations:
\ding{182} different failure types pose different effects on the DNNs pre-trained on the original training dataset, \ie, $\mathcal{D}^{\text{t}}$. For example, the accuracy of AllConvNet on the original testing dataset is around 93.0\% while the JPEG and GN reduce the results to around 80.0\% and 43.0\%, respectively, hinting that we should repair DNNs by considering the difference across different failure types. 
A simple way is to use collected failure cases that may be degraded by an unknown failure type as the guidance for repairing. 
However, in real-world applications, it is hard to collect a large number of failure cases under the similar unknown failure.
Hence, it is significantly important but difficult to explore an effective method that is able to repair the DNNs against an unknown failure type according to a few collected failure cases.
In this paper, we propose a novel DNN repairing method to address this problem and validate the effectiveness in the following experiments.
\ding{183} the accuracy reductions on the same DNN have great diversity under different failures. For example, with the Gaussian noise, the accuracy reduction is around 50\% while the value remains almost unchanged with the Brightness failure, indicating that the DNN may have very different results on different failures. Thus, the DNN repairing guided by a kind of failure should not affect the performance on other failure types and even the original clean images. 

\subsection{RQ2. Does the proposed method outperform SOTA data-driven repairing methods on the examples with specific failure patterns?}
\label{subsec:rq2}

For the $k$th failure type, we repair the AllConvNet, DenseNet, and WideResNet trained in Sec.~\ref{subsec:rq1} via the six baseline methods and the proposed method. Then, we evaluate their performance by calculating the accuracy of repaired DNNs on the failure cases, \ie, $\mathcal{D}^{\text{e}_k}$ and show the results on 15 failure types in Table~\ref{tab:cmp_sota}. In general, our method (\ie, DeepRepair) shows significant advantages over all baseline methods on the three CNN architectures under 15 failure types, demonstrating the effectiveness and generalization of the proposed method. 

Specifically, comparing with the state-of-the-art DNN repairing methods (\ie, Few-shot and SENSEI), DeepRepair achieves much higher accuracy on all three DNNs under 15 failures. In particular, on the glass blur (GB), motion blur (MB), and zoom (ZM), DeepRepair has over 500\% relative improvements on SENSEI, demonstrating the effectiveness and advantages of our method. In terms of other data augmentation-based methods, the results on the AllConvNet present that DeepRepair has much higher accuracy than all other augmentation methods under all 15 failure types. Even the state-of-the-art AugMix method still has huge accuracy gaps to our method. Nevertheless, as the DNN becomes more powerful (\ie, from AllConvNet to WideResNet), the capability of AugMix on repairing is significantly enhanced and its accuracy under several failure types (\eg, pixelate, snow, impulse noise, defocus blur, motion blur, zoom, etc.) is slightly larger than our method, indicating that AugMix is more suitable for repairing elaborately designed DNN architectures while our method obtains consistent effectiveness on the three kinds of DNNs.

\begin{table*}
    \centering
    \small
    \resizebox{1\linewidth}{!}{
    \begin{tabular}{c|c|ccccccccccccccc}
    \toprule
        \multicolumn{2}{c|}{Repair Method} & GN & SN & IN & DB & GB & MB & ZM & SW & FT & FG & BS & CT & ET & PIX & JPEG \\
    \midrule
        \multirow{7}{*}{\rotatebox{90}{AllConvNet}} 
        & Cutout		& 6.50 & 5.99 & 17.95 & 16.19 & 10.01 & 9.85 & 7.13 & 17.31 & 10.25 & 14.79 & 25.71 & 10.40 & 19.97 & 13.49 & 16.98 \\
        & Mixup			& 1.26 & 17.51 & 3.15 & 10.41 & 7.42 & 8.75 & 8.47 & 9.68 & 14.08 & 12.97 & 15.66 & 9.35 & 8.12 & 9.44 & 9.24 \\
        & CutMix		& 12.80 & 5.90 & 12.86 & 12.46 & 8.70 & 8.75 & 8.94 & 15.69 & 12.46 & 11.41 & 15.91 & 7.29 & 8.24 & 14.71 & 20.40 \\
        & AugMix		& 36.19 & 32.40 & 41.23 & 47.91 & 40.74 & 39.55 & 48.13 & 38.51 & 41.84 & 29.86 & 33.26 & 29.88 & 34.80 & 44.56 & 40.36 \\
        \cmidrule{3-17}
        & Few-shot		& 15.65 & 19.41 & 13.48 & 11.00 & 9.84 & 11.42 & 8.90 & 17.73 & 16.68 & 17.39 & 15.19 & 12.08 & 13.31 & 16.56 & 16.10 \\
        & SENSEI		& 18.85 & 25.13 & 14.27 & 11.67 & 7.29 & 9.02 & 8.41 & 15.21 & 13.18 & 15.72 & 20.78 & 9.55 & 13.72 & 16.05 & 16.18 \\
        & DeepRepair    	& \cellcolor{top1} 55.19 & \cellcolor{top1} 61.32 & \cellcolor{top1} 50.98 & \cellcolor{top1} 68.98 & \cellcolor{top1} 56.91 & \cellcolor{top1} 61.97 & \cellcolor{top1} 67.34 & \cellcolor{top1} 59.70 & \cellcolor{top1} 63.89 & \cellcolor{top1} 49.08 & \cellcolor{top1} 44.74 & \cellcolor{top1} 34.90 & \cellcolor{top1} 51.03 & \cellcolor{top1} 55.69 & \cellcolor{top1} 46.14 \\
        \midrule
        \multirow{7}{*}{\rotatebox{90}{DenseNet}} 
        & Cutout		& 11.58 & 12.73 & 23.25 & 13.25 & 8.42 & 13.57 & 10.14 & 25.71 & 16.45 & 16.54 & 39.92 & 13.72 & 27.16 & 17.13 & 24.85 \\
        & Mixup			& 14.72 & 17.13 & 3.97 & 8.17 & 3.48 & 7.19 & 6.04 & 6.90 & 12.14 & 9.63 & 5.53 & 6.31 & 11.30 & 12.60 & 11.18 \\
        & CutMix		& 15.08 & 18.12 & 17.86 & 8.65 & 11.31 & 10.55 & 6.05 & 14.98 & 10.96 & 13.24 & 17.99 & 7.95 & 11.83 & 6.56 & 16.75 \\
        & AugMix		& 48.18 & 53.62 & 57.66 & 64.63 & 52.94 & 60.61 & 63.92 & 57.23 & 62.11 & 48.37 & 50.90 & 58.36 & 55.29 & \cellcolor{top1} 65.08 & 54.71 \\
        \cmidrule{3-17} 
        & Few-shot		& 13.34 & 12.94 & 11.65 & 12.18 & 10.24 & 9.75 & 11.98 & 14.58 & 13.01 & 15.63 & 16.94 & 10.09 & 12.20 & 14.72 & 15.26 \\
        & SENSEI		& 21.24 & 16.54 & 17.28 & 15.50 & 11.94 & 7.39 & 10.31 & 16.11 & 16.08 & 14.34 & 17.17 & 7.52 & 14.33 & 17.39 & 18.78 \\
        & DeepRepair    	& \cellcolor{top1} 61.51 & \cellcolor{top1} 64.16 & \cellcolor{top1} 57.93 & \cellcolor{top1} 71.97 & \cellcolor{top1} 61.40 & \cellcolor{top1} 65.01 & \cellcolor{top1} 73.14 & \cellcolor{top1} 67.73 & \cellcolor{top1} 64.00 & \cellcolor{top1} 57.81 & \cellcolor{top1} 56.94 & \cellcolor{top1} 67.00 & \cellcolor{top1} 62.56 & 60.32 & \cellcolor{top1} 63.41 \\
        \midrule
        \multirow{7}{*}{\rotatebox{90}{WideResNet}} 
        & Cutout		& 13.06 & 13.17 & 25.02 & 14.95 & 10.28 & 12.48 & 12.87 & 26.45 & 15.25 & 22.53 & 39.23 & 16.13 & 21.23 & 20.70 & 29.00 \\
        & Mixup			& 11.57 & 13.66 & 11.21 & 7.44 & 3.50 & 5.71 & 16.12 & 6.21 & 7.88 & 7.48 & 13.11 & 4.68 & 6.04 & 5.65 & 8.64 \\
        & CutMix		& 11.43 & 10.62 & 19.20 & 10.70 & 6.71 & 6.56 & 5.84 & 14.50 & 10.10 & 11.55 & 19.07 & 14.19 & 10.30 & 13.10 & 12.98 \\
        & AugMix		& 65.59 & \cellcolor{top1} 69.73 & \cellcolor{top1} 71.57 & \cellcolor{top1} 79.36 & 65.95 & \cellcolor{top1} 77.72 & \cellcolor{top1} 81.17 & 68.77 & 69.99 & 62.09 & \cellcolor{top1} 65.65 & \cellcolor{top1} 69.70 & \cellcolor{top1} 68.97 & 63.44 & \cellcolor{top1} 65.64 \\
        \cmidrule{3-17} 
        & Few-shot		& 12.52 & 15.13 & 11.10 & 10.85 & 13.28 & 10.32 & 12.54 & 10.28 & 12.73 & 12.36 & 14.04 & 6.20 & 12.83 & 12.21 & 12.62 \\
        & SENSEI		& 10.60 & 5.21 & 4.75 & 14.49 & 8.27 & 10.38 & 12.40 & 9.47 & 7.33 & 10.00 & 10.74 & 10.68 & 17.27 & 13.96 & 18.96 \\
        & DeepRepair        & \cellcolor{top1} 67.80 & 69.44 & 71.10 & 74.99 & \cellcolor{top1} 66.41 & 77.33 & 80.42 & \cellcolor{top1} 71.26 & \cellcolor{top1} 74.13 & \cellcolor{top1} 67.09 & 62.37 & 64.54 & 67.10 & \cellcolor{top1} 67.73 & 65.03 \\
    \bottomrule
    \end{tabular}
    }
    \caption{Accuracy of  repaired DNN architectures on 15 failure datasets, \ie, $\{\mathcal{D}^{\text{e}_k}|k=[1,15]\}$. We choose CutMix, Mixup, Cutout, Few-Shot, SENSEI, AugMix as baseline methods, and fine-tune them with DeepRepair under 15 failure patterns, respectively. We highlight the best results with red.}
    \label{tab:cmp_sota}
\end{table*}

\subsection{RQ3. Does the proposed method harm the robustness to other failure patterns and the accuracy on clean images?}

A well repaired DNN should achieve much higher accuracy on the target failure patterns while not harming the accuracy on the original clean images and the robustness to other kinds of patterns.

\begin{figure}[t]
    \centering
    \includegraphics[width=\linewidth]{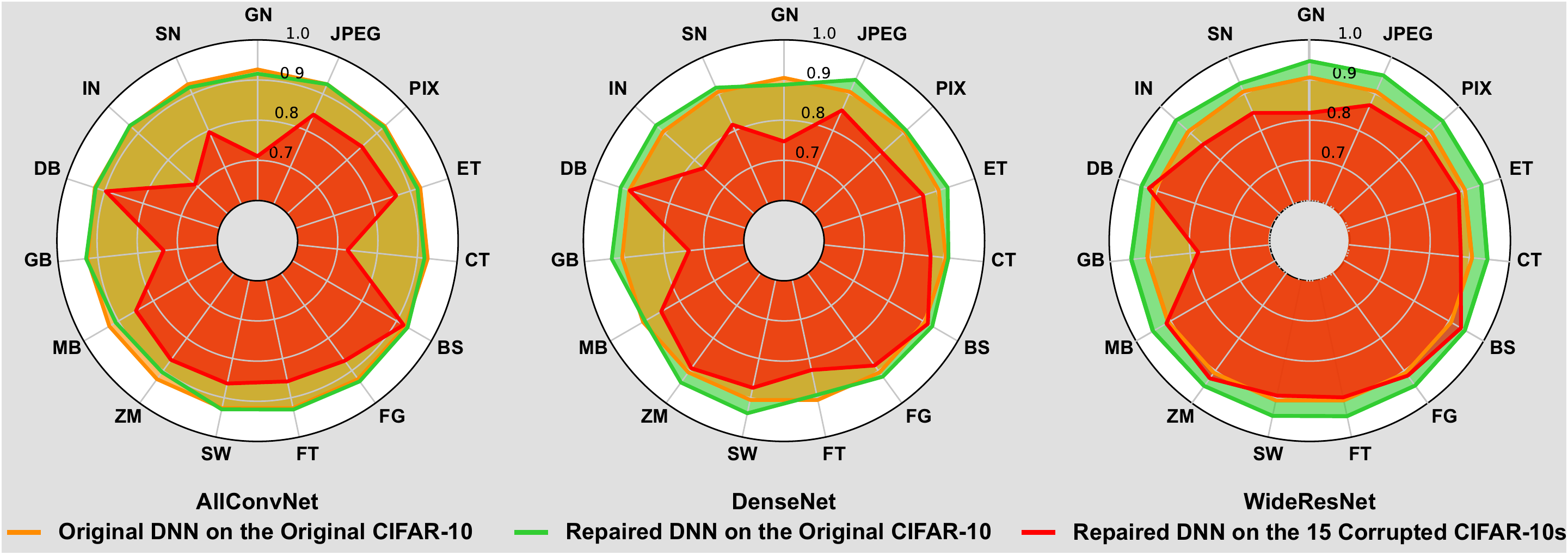}
    \caption{Accuracy of original and repaired DNNs on the original CIFAR-10's testing dataset (\ie, $\mathcal{D}^\text{v}$) and 15 extended testing datasets (\ie, $\mathcal{D}^{\text{v}_k}$), respectively.}
    \label{fig:rq3_original_repaired}
\end{figure}

To validate the first capability of our method, we evaluate the repaired DNN under the $k$th failure pattern on the original testing dataset (\ie, $\mathcal{D}^{\text{v}}$) and compare with the original DNN. We present the results in Fig.~\ref{fig:rq3_original_repaired}. Specifically, for the $k$th axes, we show the evaluation results of the original and the repaired DNN on the $\mathcal{D}^\text{v}$ and the accuracy of repaired DNN on the $\mathcal{D}^{\text{v}_k}$. We have the following observations:
\ding{182} Comparing the accuracy difference of DNNs on the original testing dataset (\ie, green line for the repaired DNN and yellow line for the original DNN) and the unknown failure datasets (\ie, red line for the repaired DNN and black line for the original DNN), we observe that the accuracy difference has significantly decreased after repairing, demonstrating that the proposed method could effectively repair the original DNN.
\ding{183} Comparing the results on the original testing dataset  $\mathcal{D}^\text{v}$ (\ie, the yellow line and the green line), we see that all repaired DNNs do not harm the accuracy of the original DNN on the clean images and even achieve much higher accuracy when we repair the DenseNet and WideResNet.

\begin{figure*}
    \centering
    \includegraphics[width=\linewidth]{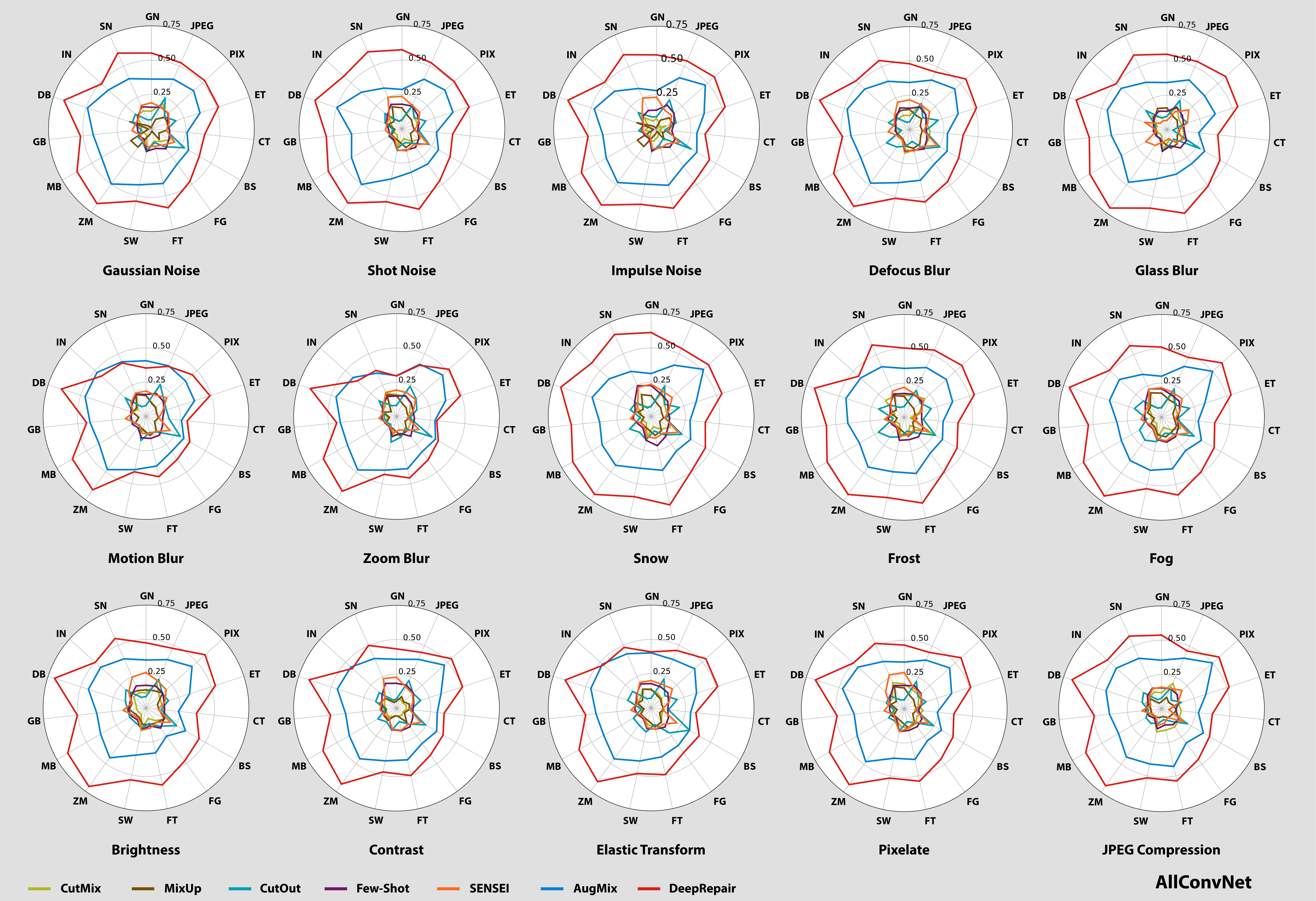}\\
    \caption{Comparing the repairing methods on AllConvNet by evaluating the accuracy of repaired DNN under one failure pattern (\ie, the name at the bottom of each sub-figure) on other failure datasets, \ie, $\{\mathcal{D}^{\text{e}_k}|k=[1,15]\}$.}
    \label{fig:rq3_robust}
\end{figure*}

To validate the second capability of our method, we take the AllConvNet as an example and evaluate the accuracy of the repaired DNN based on one failure pattern on other failure datasets to show whether the repaired model could have higher robustness on other failure patterns or not. As shown in Fig.~\ref{fig:rq3_robust}, DeepRepair makes the repaired DNN under one failure pattern achieve similar significant accuracy enhancement on other failure datasets, outperforming all baseline methods.
Overall, the two experiments in Fig.~\ref{fig:rq3_original_repaired} and \ref{fig:rq3_robust} demonstrate that DeepRepair can effectively repair the DNN while not harming the accuracy on the clean dataset as well as the robustness on other failure pattern-based datasets. 

\begin{table*}[ht!]
    \centering
    \small
    \resizebox{1\linewidth}{!}{
    \begin{tabular}{c|c|ccccccccccccccc}
    \toprule
        \multicolumn{2}{c|}{Repair Method} & GN & SN & IN & DB & GB & MB & ZM & SW & FT & FG & BS & CT & ET & PIX & JPEG \\
    \midrule
        \multirow{3}{*}{\rotatebox{90}{AllConv}} 
        & Original      & 0 & 0 & 0 & 0 & 0 & 0 & 0 & 0 & 0 & 0 & 0 & 0 & 0 & 0 & 0 \\
        & no-cluster 	& \cellcolor{top1} 58.13 & 61.27 & 48.47 & \cellcolor{top1} 69.59 & \cellcolor{top1} 57.35 & 54.95 & 59.17 & 59.28 & 63.78 & \cellcolor{top1} 49.91 & \cellcolor{top1} 44.85 & 33.82 & 47.57 & 52.57 & 45.73 \\
        & cluster 		& 55.19 & \cellcolor{top1} 61.32 & \cellcolor{top1} 50.98 & 68.98 & 56.91 & \cellcolor{top1} 61.97 & \cellcolor{top1} 67.34 & \cellcolor{top1} 59.70 & \cellcolor{top1} 63.89 & 49.08 & 44.74 & \cellcolor{top1} 34.90 & \cellcolor{top1} 51.03 & \cellcolor{top1} 55.69 & \cellcolor{top1} 46.14 \\
    \midrule
        \multirow{3}{*}{\rotatebox{90}{Dense}} 
        & Original      & 0 & 0 & 0 & 0 & 0 & 0 & 0 & 0 & 0 & 0 & 0 & 0 & 0 & 0 & 0 \\
        & no-cluster 	& 43.93 & 61.10 & \cellcolor{top1} 58.66 & \cellcolor{top1} 72.96 & 60.54 & 62.05 & \cellcolor{top1} 74.49 & 67.34 & 61.59 & 55.97 & 55.78 & 63.64 & 61.42 & \cellcolor{top1} 65.87 & 63.18 \\
        & cluster 		& \cellcolor{top1} 61.51 & \cellcolor{top1} 64.16 & 57.93 & 71.97 & \cellcolor{top1} 61.40 & \cellcolor{top1} 65.01 & 73.14 & \cellcolor{top1} 67.73 & \cellcolor{top1} 64.00 & \cellcolor{top1} 57.81 & \cellcolor{top1} 56.94 & \cellcolor{top1} 67.00 & \cellcolor{top1} 62.56 & 60.32 & \cellcolor{top1} 63.41 \\
    \midrule
        \multirow{3}{*}{\rotatebox{90}{WideRes}} 
        & Original      & 0 & 0 & 0 & 0 & 0 & 0 & 0 & 0 & 0 & 0 & 0 & 0 & 0 & 0 & 0 \\
        & no-cluster 	& 62.72 & 50.80 & 65.61 & \cellcolor{top1} 76.13 & 64.53 & 75.46 & 79.16 & 69.47 & 70.18 & 64.83 & 61.37 & \cellcolor{top1} 66.45 & 66.88 & 61.23 & \cellcolor{top1} 65.21 \\
        & cluster 	    & \cellcolor{top1} 67.80 & \cellcolor{top1} 69.44 & \cellcolor{top1} 71.10 & 74.99 & \cellcolor{top1} 66.41 & \cellcolor{top1} 77.33 & \cellcolor{top1} 80.42 & \cellcolor{top1} 71.26 & \cellcolor{top1} 74.13 & \cellcolor{top1} 67.09 & \cellcolor{top1} 62.37 & 64.54 & \cellcolor{top1} 67.10 & \cellcolor{top1} 67.73 & 65.03 \\
    \bottomrule
    \end{tabular}
    }
    \caption{Ablation study of DeepRepair.We consider two variants. The first use uniform sampling to select reference images for style-guided data augmentation (\ie, Eq.~\ref{eq:stypleop}) and we denote it as `no-cluster'. The second one use Eq.~\ref{eq:clsstypleop} for clustering-guided sampling and we denote it as `cluster'. We compare the two methods via the accuracy of repaired DNNs on $\{\mathcal{D}^{\text{e}_k}|k=[1,15]\}$. We highlight the best result with red. }
    \label{tab:ablation}
\end{table*}

\begin{figure*}
    \centering
    \includegraphics[width=\linewidth]{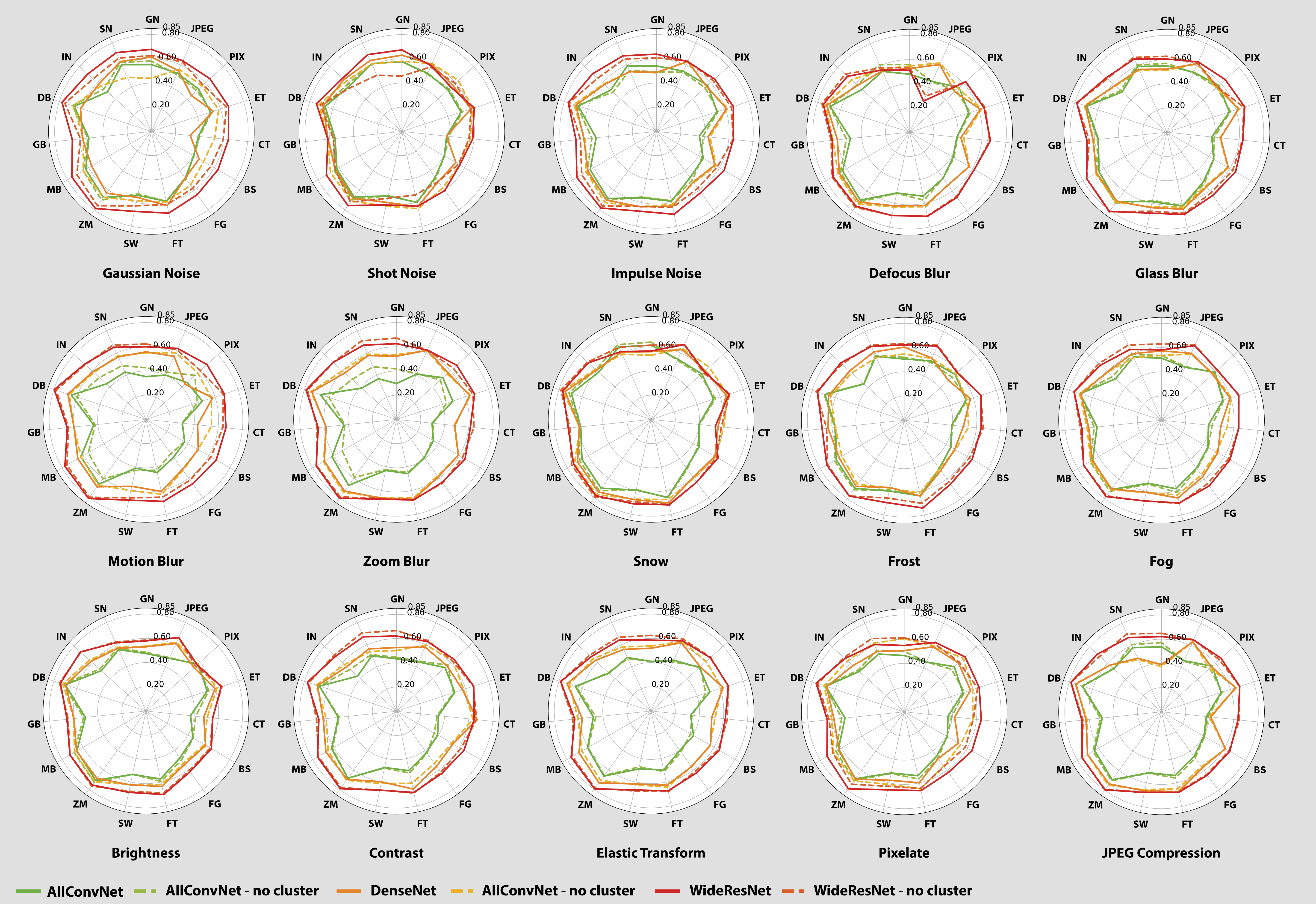}\\
    \caption{Comparing the two variants of our methods on three DNNs by evaluating the accuracy of repaired DNN under one failure pattern (\ie, the name at the bottom of each sub-figure) on other failure datasets, \ie, $\{\mathcal{D}^{\text{e}_k}|k=[1,15]\}$.}
    \label{fig:ablation}
\end{figure*}

\subsection{RQ4. Do the proposed components all contribute the final accuracy?}

To demonstrate the effectiveness of our clustering-based style-guided data augmentation for DNN repairing, we conduct an ablation study by repairing pre-trained AllConv, Dense, and WideResNet models with two variants of our method.
The first one performs the style transfer on randomly selected failure cases to guide the repairing process without using the clustering method and we denote this variant as `no-cluster' in Table~\ref{tab:ablation} and Fig.~\ref{fig:ablation}. The second one is our final version while the clustering is first done on the collected failure cases and we call this variant as `cluster' in Table~\ref{tab:ablation} and Fig.~\ref{fig:ablation}.
According to the reported results, we have the following observations: \ding{182} As shown in Table~\ref{tab:ablation}, both variants enhance the accuracy on all failure patterns significantly, demonstrating that the proposed style-guided data augmentation indeed can repair the DNN under some specific patterns effectively. \ding{183} Comparing the accuracy of repaired DNNs based on `no-cluster' and `cluster', our final version with the clustering method outperforms the `no-cluster' one under most of the failure patterns on all three DNNs, demonstrating that the proposed clustering-based style-guided data augmentation does help enhance the robustness against various failure patterns. \ding{184} In Fig.~\ref{fig:ablation}, we further evaluate the accuracy of repaired DNNs under one failure pattern on other failure datasets to show whether the clustering harms the accuracy of repaired DNNs on other failure patterns. As shown in Fig.~\ref{fig:ablation}, the repaired DNNs under one failure pattern based on clustering-based variant also achieve significant accuracy on other patterns and usually show better accuracy than the repaired DNNs based on `no-cluster' variant, demonstrating that the clustering is able to enhance the accuracy of repaired DNNs on other failure patterns further.

\section{Conclusion}\label{sec:concl}


In this paper, we tackle the imminent DNN repairing problem towards the real-world operational environment. To address the issue that there exists a mismatch between the distributions of the training dataset and the real-world testing data that may be corrupted by unknown factors in the operational environment such as weather elements, blur, noise, \etc., we resort to a style-guided data augmentation paradigm that not only bridges the aforementioned distributional gap, but also can retain high DNN performance while handling normal or clean data. Specifically, we first identify that the data augmentation-based DNN repairing solution can help to address the problem and avoid the over-fitting risk, which, however, raises a new challenge, \ie, how to introduce the failure pattern into the data augmentation process with a limited small number of collected failure example data. To further address this challenge, we then propose the \textit{style-guided data augmentation for DNN repairing} where a style transfer is used to introduce the unknown failure patterns within the failure data into the training data via data augmentation. Moreover, we propose the \textit{clustering-based corrupted data generation} for much more effective style-guided data augmentation. We have conducted a set of comprehensive experiments with fifteen degradation factors that may happen in the real world and compare with four state-of-the-art data augmentation methods and two DNN repairing methods, demonstrating that our method is able to significantly enhance the deployed DNNs on the failure data with even better accuracy on clean datasets. 

We have demonstrated that our proposed DNN repairing method can handle the real-world failure patterns that are naturally occurred very well. Going beyond, one of the future research directions can be to extend our \emph{DeepRepair} capability towards handling perturbations that are not just naturally occurred, but adversarially or maliciously generated by an adversary. These adversarially crafted perturbations, ranging from the most common additive noise-based ones \cite{madry2017towards,goodfellow2014explaining,papernot2017practical,guo2020spark}, to non-additive noise-based ones such as blur effect \cite{neurips20_abba}, denoise effect \cite{arxiv20_pasadena}, geometric morphing \cite{acmmm20_amora}, exposure effect \cite{cheng2020adversarial,tian2020bias}, rain effect \cite{zhai2020s,guo2020efficientderain}, \etc, are usually very stealthy and imperceptible, thus posing yet another challenge for DNN-based software repairing.

\balance

\bibliographystyle{IEEEtran}
\bibliography{ref}

%








\end{document}